# An Algebraic Semantics for Possibilistic Logic

**Luca Boldrin**
Dept. of Pure and Applied Math.
University of Padova
Via Belzoni 7 35100 Padova PD, Italy
boldrin@ladseb.pd.cnr.it

**Claudio Sossai**
Ladseb-CNR
Corso Stati Uniti 4
35127 Padova PD, Italy
sossai@ladseb.pd.cnr.it

**Abstract**

The first contribution of this paper is the presentation of a Pavelka–like formulation of possibilistic logic in which the language is naturally enriched by two connectives which represent negation ($\neg$) and a new type of conjunction ($\otimes$). The space of truth values for this logic is the lattice of possibility functions, that, from an algebraic point of view, forms a *quantal*. A second contribution comes from the understanding of the new conjunction as the combination of tokens of information coming from different sources, which makes our language "dynamic". A Gentzen calculus is presented, which is proved sound and complete with respect to the given semantics. The problem of truth functionality is discussed in this context.

## 1 INTRODUCTION

The distinction between truth-functional and non truth-functional logics have been widely stressed by several authors [Dubois, Lang and Prade 91], [Hajek et. al 94]. The first class contains many valued logics in the style of Łukasiewicz, and has been given a precise status since the work of Goguen and Pavelka (see [Pavelka 79]); while Pavelka proposed a general frame in which the set of truth values was a complete lattice ordered monoid, he limited his analysis to finite chains and to the unit interval of reals.

The second class contains measure-based logics, where the underlying measure can have different natures: a probability, a possibility, a belief function, etc. (see, for instance, [Fagin and Halpern 94], [Dubois, Lang and Prade 94], [Saffiotti 92]). It has been given a formal status in terms of modal logics in [Hajek et. al 94][1] and [Voorbraak 93]. Links to classical epistemic logics have been established in [Murai et al. 93].

In the present paper we show that possibilistic logic can be embedded in a many-valued (truth-functional) logic, where the set of truth values is not the unit interval of real numbers, but a complete lattice made of possibility distributions (more precisely, a *quantal*)[2]. Moreover, as in Pavelka and in [Takeuti and Titani 84], we introduce in the language a suitable subset of the truth values; we show that the resulting logic, which fits in fact in Pavelka's general frame, contains possibilistic logic as expressed in [Dubois, Lang and Prade 94].

In this many-valued view the truth value of a formula is the set of possibility distributions which satisfy it (in the usual sense). Composition of formulae through logical connectives in the language has a semantical counterpart in the corresponding composition of these sets of possibility distributions. The trick of introducing truth values into the language, token from Pavelka, is used in this context to account for the measure values from inside the language.

A second point of concern in our work is the *dynamics* of information. In the field of measure-based logics, it seems that most of the languages are *static*, in the sense that they perform inference on a unique information state; no logical counterpart has been established to the combination of evidence which, semantically, is a change of information state[3]. In modal formulations the $\wedge$ operator between modal formulae like $N_\alpha A$ can, in fact, be thought of as a connective representing *expansion*, which is a change (more precisely, a refinement) of the information state; however, no other dynamic connectives are available.

---

[1] As a matter of facts Hajek does more than that, since he puts together a truth-functional and a measure based logic in the same framework.

[2] It is well known that it is not possible to make a compositional classical logic whose truth space is the unit interval, since the unit interval cannot be imposed a Boolean structure (it can however be given a Complete Heyting Algebra structure, so that it works as the truth value space of an intuitionistic logic, see [Takeuti and Titani 84]).

[3] This is not completely true with Saffiotti's Belief Function Logic, since the conjunction of BF-formulae corresponds in some way to Dempster's combination on models

A parallel can be drawn with the representation of certain knowledge; in that case, since there is no way of weighing evidence from different sources, the only meaningful dynamic operators correspond to expansion, revision or update. An interesting work in the direction of representing these operators as logical connectives is [de Rijke 94] which seeks to capture the non-determinism of revision. The uncertain setting has to cope with a much wider set of operators, and we believe that it would be of a certain interest to explore their representation *inside* the language. For instance, it would be interesting to give a syntactical account of the Dempster-Shafer combination operator for belief functions (a syntactic characterization of Dempster conditionalization has been proposed in [Alechina and Smets 94]), or of the Jeffrey conditionalization operator for probabilities; in this paper we start this work from an easier task, which is the syntactical representation of the Lukasiewicz combination operator in the possibilistic framework.

Let us assume that a piece of evidence is modeled by a possibility distribution on a set of possible worlds; as reported in [Dubois and Prade 85], there are several ways of aggregating information, depending on the relation among the sources of information, and on assumptions on their reliability. In this paper we focus on two operators for combining possibility distributions: $\wedge$ defined by $(\pi_1 \wedge \pi_2)(w) = \pi_1(w) \wedge \pi_2(w)$ and the Lukasiewicz operator $\times$, defined by $(\pi_1 \times \pi_2)(w) = (\pi_1(w) + \pi_2(w) - 1) \vee 0$; both of these operators are T-norms. The first one, up to when the result remains consistent, can be used to model expansion, i.e. the combination of coherent information (think for instance of tokens of information coming from the same source): we represent it syntactically by "&"; it corresponds to the set union of possibilistic formulae in the logic of [Dubois, Lang and Prade 94], or to the $\wedge$ between modal formulae in modal approaches. The second operator models combination of evidence coming from distinct sources: if the two evidences agree, combination yields an evidence stronger than both[4]. We introduce it in the language as "$\otimes$", which has no explicit representation in standard possibilistic logic. Moreover, it is very natural to add to the language a negation, which corresponds to the operation of fuzzy set complementation with respect to $\times$. This approach, while owing much to modal approaches, as it can be argued from the semantics, differs from them in that it explicitly introduces numerical constants in the language (which, as we discussed above, are just some of the truth values), and in the semantics of negation, which we are going to discuss in some detail later. For the time being it suffice to say that modal negation deals with absence of information in some information state, while our negation represents actual disbelief.

The result is a logic endowed with a sound and complete Gentzen calculus; we named it Dynamic Possibilistic Logic because of the dynamic interpretation we just gave. To improve the clarity of the paper, we chose to introduce the propositional system (DPL) first (par. 2), and to extended it to the full predicative system (DPL*) in a second step (par 3). The reader could notice that our approach shares some features with Lehman's plausibility logic [Lehman 91]; the substantial difference is that Lehman seeks to capture non-monotonicity, and so plausibility logic enjoys contraction but not weakening and full cut. On the contrary, our connective $\otimes$ is monotonic but not idempotent, since it is meant to represent combination of information from different sources; consequently, our logic misses out contraction but allows for cut and weakening. As a matter of fact, the logic happens to fall in the field of substructural logics, since it can be seen as a specialization of Girard's *linear logic* [Girard 87]. This is not surprising, in the light of the informational interpretation of substructural logics pointed out in [Wansing 93].

## 2  THE PROPOSITIONAL SYSTEM

### 2.1  THE SEMANTICS

We assume the following language, where $\alpha$ for any $\alpha \in [0,1]$ are constants. The set of atomic propositions is named $\mathcal{L}_0$.

formula ::= atomic_proposition | $\alpha$ | $\neg$ formula | formula & formula | formula $\otimes$ formula

We take $\mathcal{L}$ to be the set of formulae; it is convenient to define $\mathcal{L}_1$ as the set of formulae with no occurrences of $\alpha$ constants for any $\alpha \in (0,1)$ — notice that **0** and **1** are in $\mathcal{L}_1$. We use upper case Latin letters (A, B, C,...) for formulae, while reserving L, M, N for $\mathcal{L}_1$-formulae, and upper case greek letters ($\Gamma$, $\Delta$, ...) for multisets of formulae; the greek letters $\alpha$ and $\beta$ always represent real numbers in $[0,1]$.

We introduce some new symbols via the definitions:

$$A \rightarrow B =_{def} \neg(A \otimes \neg B)$$
$$A \oplus B =_{def} \neg(\neg A \& \neg B)$$
$$A \mathbin{\invamp} B =_{def} \neg(\neg A \otimes \neg B)$$

The language is the same as in [Pavelka 79], where our & corresponds to $\wedge$ and our $\otimes$ to $\oplus$. Our choice of the connectives differs from Pavelka's, since we want to stress the proximity of our logic to substructural logics in the style of [Girard 87].

Let $\mathcal{P}$ denote the set of functions (which we call possibility distributions) from a non-empty set $W$ to the real interval $[0,1]$, with the order $\leq$ ($\pi_1 \leq \pi_2$ iff for any $w$ it holds that $\pi_1(w) \leq \pi_2(w)$); the lattice operations $\vee$ and $\wedge$ on possibility functions are defined with respect to the order $\leq$; $\langle P, \vee, \wedge \rangle$ is a complete lattice. The operation $\times$ is defined by $\pi = \pi_1 \times \pi_2$ iff for any $w$  $\pi(w) = 0 \vee (\pi_1(w) + \pi_2(w) - 1))$. To define the semantics of negation, we make use of fuzzy set complementation with respect to $\times$, which makes our negation coincide with Girard's, where the inconsistent set contains only the function identically zero.

---

[4]this role can also be played by the product $\cdot$ defined by $(\pi_1 \cdot \pi_2)(w) = \pi_1(w) \cdot \pi_2(w)$; in [Dubois and Prade 85] the authors claim that this operator models the combination of information from distinct and independent sources. The syntactical representation of this operator has been studied in [Boldrin 94].

We need the following definitions:

**Def. 1**  *1. For any $\pi$, $\downarrow \pi = \{\sigma \in \mathcal{P} : \sigma \leq \pi\}$.*

*2. For any $\alpha \in [0,1]$, $\boldsymbol{\alpha}$ is the function identically equal to $\alpha$ (in particular $\mathbf{1}(w) = 1$ and $\mathbf{0}(w) = 0$ for any $w$).*

*3. For any $G \subseteq \mathcal{P}$ and $H \subseteq \mathcal{P}$, $G \Rightarrow H = \{\sigma : \forall \pi \, (\pi \in G \to \pi \times \sigma \in H)\}$.*

*4. For any $G \subseteq \mathcal{P}$, $G^\perp = G \Rightarrow \{\mathbf{0}\}$.*

It can be easily verified that $^{\perp\perp}$ is a closure operator on $2^\mathcal{P}$ (see [Girard 87]).

**Theorem 1** *For any $G \subseteq \mathcal{P}$, let $\pi_G = \bigvee_{\pi \in G} \pi$; then:*

*1. $G^\perp = \{\pi : \pi \times \pi_G = \mathbf{0}\}$*

*2. $G^{\perp\perp} = \downarrow \pi_G$*

The structure $\langle \mathcal{P}, \times, \mathbf{1} \rangle$ is a commutative monoid with unit, and $\perp = \{\mathbf{0}\} \subseteq \mathcal{P}$. Hence the structure $\langle \mathcal{P}, \times, \mathbf{1}, \perp \rangle$ is a phase space in Girard's sense. The closure operator is exactly the one of Girard, so the set $\mathcal{Q} = \{G \subseteq \mathcal{P} : G = G^{\perp\perp}\}$ is the set of *facts*, and belongs to the class of *Girard quantales* as defined in [Rosenthal 90].

A frame for our language is a couple: $F = \langle W, V_0 \rangle$, where W is a nonempty set of worlds, $V_0 : \mathcal{L}_0 \to 2^W$ is a propositional assignment over the worlds which is extended to $V : \mathcal{L}_1 \to 2^W$, as usual.

**Def. 2** *Given the frame $F$, let us define the function $\| \cdot \|_F : \mathcal{L} \to \mathcal{Q}$:*

$$\|p\|_F =_{\text{def}} \{\pi : Nec_\pi(V(p)) = 1\}$$
$$\|\boldsymbol{\alpha}\|_F =_{\text{def}} \downarrow \boldsymbol{\alpha}$$
$$\|\neg A\|_F =_{\text{def}} \|A\|_F^\perp$$
$$\|A \,\&\, B\|_F =_{\text{def}} \|A\|_F \cap \|B\|_F$$
$$\|A \otimes B\|_F =_{\text{def}} \|A\|_F \times \|B\|_F$$

where $Nec_\pi : 2^w \to [0,1]$ is the necessity function associated to the possibility distribution $\pi$: $Nec_\pi(X) = 1 - \bigvee_{w \notin X} \pi(w)$. The $\times$ product between sets is the point to point product.

It can be verified that :

$$\|A \to B\|_F = \|A\|_F \Rightarrow \|B\|_F$$
$$\|A \oplus B\|_F = \|A\|_F \vee \|B\|_F$$
$$\|A \,\mathfrak{P}\, B\|_F = (\|A\|_F + \|B\|_F) \wedge \mathbf{1}$$

$$\|L\|_F = \{\pi : Nec_\pi(V(L)) = 1\} \text{ for any } L \in \mathcal{L}_1$$
$$\|\alpha \to L\|_F = \{\pi : Nec_\pi(V(L)) \geq \alpha\} \text{ for any } L \in \mathcal{L}_1$$

It follows from theorem 2 that $\|A\|$ as above defined is a fact (i.e. belongs to $\mathcal{Q}$) for any $A \in \mathcal{L}$. It is worth noting that the subset $\mathcal{B} = \{\pi : (\forall w \in W)(\pi(w) \in \{0,1\})\}$ is a Boolean algebra contained in $\mathcal{Q}$. For this reason, $\mathcal{L}_1$-formulae (whose value is in $\mathcal{B}$ — see the fourth equivalence above) behave classically.

The fifth equivalence is very important, since it states that the formula $\boldsymbol{\alpha} \to L$ has the same meaning as the possibilistic formula $(L, \alpha)$ in [Dubois, Lang and Prade 94], or as the formula in modal flavour $N_\alpha L$ in the style of Hajek.

**Def. 3** *A model is a couple $K = \langle F, \pi \rangle$ where $F$ is a frame. We say that $K \models A$ iff $\pi \in \|A\|_F$. A formula $A$ is valid in $F$ iff for any model $K$ in the frame $F$, $K \models A$ or, equivalently (see lemma 1 in paragraph 4) iff $\mathbf{1} \in \|A\|_F$.*

In a fixed frame a formula $\alpha \to A$ is true in the models whose possibility distribution gives $A$ at least $\alpha$ support; a formula $\neg A$ is true in the models which are inconsistent with the models for $A$; a formula $A \,\&\, B$ is true in those models which fit both $A$ and $B$; and, eventually, a formula $A \otimes B$ is true in any model whose possibility distribution is the product of one of an $A$-model and one of a $B$-model. Since the lattice $\mathcal{P}$ is complete, we can establish a correspondence $U_F : \mathcal{L} \to \mathcal{P}$ between formulae and their least informative model in a frame (keep in mind that least informative means higher in the order $\leq$):

$$U_F(A) =_{\text{def}} \bigvee_{\pi \in \|A\|_F} \pi$$

**Theorem 2** *Given a frame $F$, $\|A\|_F = \|A\|_F^{\perp\perp} = \downarrow U_F(A)$. Moreover, the following statements hold:*

*1. $U_F(\boldsymbol{\alpha}) = \boldsymbol{\alpha}$*

*2. $U_F(L) = \lambda w. \begin{cases} 1 & \text{if } w \in V(L) \\ 0 & \text{otherwise} \end{cases}$ for any $\mathcal{L}_1$-formula $L$*

*3. $U_F(\neg A) = \mathbf{1} - U_F(A)$*

*4. $U_F(A \,\&\, B) = U_F(A) \wedge U_F(B)$*

*5. $U_F(A \otimes B) = U_F(A) \times U_F(B) = (U_F(A) + U_F(B) - 1) \vee \mathbf{0}$*

*6. $U_F(A \oplus B) = U_F(A) \vee U_F(B)$*

*7. $U_F(A \,\mathfrak{P}\, B) = (U_F(A) + U_F(B)) \wedge \mathbf{1}$*

*8. $U_F(A \to B) = (\mathbf{1} - U_F(A) + U_F(B)) \wedge \mathbf{1}$*

To define the semantic entailment relation, we first consider the entailment between formulae:

$A \models B$ iff for any frame $F$, $U_F(A) \leq U_F(B)$

We can now state the following

**Theorem 3** *Let the frame $C = \langle W_c, V_0^c \rangle$ be defined as follows: $W_c$ is the set of classical propositional valuations for $\mathcal{L}_0$ (i.e. the set of functions from $\mathcal{L}_0$ to $\{True, False\}$) and $V_0^c(p) = \{w \in W_c : w \models p \text{ (classically)}\}$. Then $U_C(A) \leq U_C(B)$ implies $U_F(A) \leq U_F(B)$ for any frame $F$.*

An important consequence of the theorem above is that we can restrict our attention to a unique quantal, which is the one made from the set of possibility distributions over the set $W_c$, via the closure operation. In fact, we can define the semantic entailment relation as follows (here and in the following we write $U(A)$ for $U_C(A)$):

$$A \models B \text{ iff } U(A) \leq U(B)$$

Eventually, since the intended meaning of the sequent $\Gamma \vdash \Delta$ is $\bigotimes_{A \in \Gamma} A \to \mathfrak{N}_{B \in \Delta} B$, then we say that the sequent $\Gamma \vdash \Delta$ is valid iff $\bigotimes_{A \in \Gamma} A \models \mathfrak{N}_{B \in \Delta} B$.

Let us now briefly comment on negation. Possibilistic models on the same frame represent a state of information about the possible worlds of the frame; they are informationally ordered: $\pi_1 \leq \pi_2$ means that $\pi_1$ is more informative then $\pi_2$, since it better constrains the set of possible worlds. Since a formula is interpreted in the least informative information state which satisfies it, we have two possible readings for negation: the first is the *modal* one, which refers to information which is absent in a given information state. In this case the statement $K \models \neg N_\alpha L$ must be read as: "in the given state of information it is not possible to prove that $L$ is necessary at least $\alpha$ (while it may become possible in a refinement of the information state)". The second interpretation for negation, which we use in our logic, is an *internal* one, in the sense that the statement $K \models \neg(\alpha \to L)$ is read as: "in the given state of information we definitely refuse to accept that $L$ be necessary at least $\alpha$ (and no refinement of this information state will allow to prove the opposite)". The formula $\neg(\alpha \to L)$ then expresses an effective token of information, and does not deal with absence of information. Note that, if applied to a classical framework, this second reading would make $\neg \Box L$ equivalent to $\Box \neg L$, but this is not the case here. Moreover, the reader can verify that there are models with non-zero possibility functions which satisfy both $A$ and $\neg A$; all of these functions are, however, smaller than **0.5**. So we tolerate that a partially consistent information state can support both a token of information and its negation.

## 2.2 THE PROOF SYSTEM DPL

The proof system will be given in a Gentzen-style calculus, since it is the most comfortable way to deal with multisets (remember that, because of the absence of contraction, it does matter how many times a formula is given). Another reason for choosing this calculus is the possible generalization to cases in which other rules are not accepted (in the style of plausibility logic [Lehman 91]). The DPL calculus consists of four parts: structural rules, logical rules, an axiom for distributivity, and three further "numerical" axioms for characterizing the behaviour of constants.

**Structural rules:**

id)   $A \vdash A$   cut) $\dfrac{\Gamma \vdash B, \Delta \quad \Gamma', B \vdash \Delta'}{\Gamma, \Gamma' \vdash \Delta, \Delta'}$

exL) $\dfrac{\Gamma, B, A, \Delta \vdash \Lambda}{\Gamma, A, B, \Delta \vdash \Lambda}$   exR) $\dfrac{\Gamma \vdash \Delta, B, A, \Lambda}{\Gamma \vdash \Delta, A, B, \Lambda}$

wL) $\dfrac{\Gamma \vdash \Delta}{\Gamma, A \vdash \Delta}$   wR) $\dfrac{\Gamma \vdash \Delta}{\Gamma \vdash \Delta, A}$

abs) $\dfrac{\Gamma, B \vdash L, \Delta}{\Gamma, B \vdash L \otimes B, \Delta}$   $L \in \mathcal{L}_1$

**Logical rules:**

&) $\dfrac{\Gamma, A \vdash \Delta}{\Gamma, A \& B \vdash \Delta} \quad \dfrac{\Gamma, B \vdash \Delta}{\Gamma, A \& B \vdash \Delta}$   $\dfrac{\Gamma \vdash A, \Delta \quad \Gamma \vdash B, \Delta'}{\Gamma \vdash A \& B, \Delta, \Delta'}$

$\otimes$) $\dfrac{\Gamma, A, B \vdash \Delta}{\Gamma, A \otimes B \vdash \Delta}$   $\dfrac{\Gamma \vdash A, \Delta \quad \Gamma' \vdash B, \Delta'}{\Gamma, \Gamma' \vdash A \otimes B, \Delta, \Delta'}$

$\oplus$) $\dfrac{\Gamma, A \vdash \Delta \quad \Gamma, B \vdash \Delta}{\Gamma, B \oplus A \vdash \Delta}$   $\dfrac{\Gamma \vdash A, \Delta}{\Gamma \vdash A \oplus B, \Delta} \quad \dfrac{\Gamma \vdash B, \Delta}{\Gamma \vdash A \oplus B, \Delta}$

$\mathfrak{N}$) $\dfrac{\Gamma, A \vdash \Delta \quad \Gamma', B \vdash \Delta'}{\Gamma, \Gamma', A \mathfrak{N} B \vdash \Delta, \Delta'}$   $\dfrac{\Gamma \vdash A, B, \Delta}{\Gamma \vdash A \mathfrak{N} B, \Delta}$

$\to$) $\dfrac{\Gamma \vdash A, \Delta \quad \Gamma', B \vdash \Delta'}{\Gamma, \Gamma', A \to B \vdash \Delta, \Delta'}$   $\dfrac{\Gamma, A \vdash B, \Delta}{\Gamma \vdash A \to B, \Delta}$

$\neg$) $\dfrac{\Gamma \vdash A, \Delta}{\Gamma, \neg A \vdash \Delta}$   $\dfrac{\Gamma, A \vdash \Delta}{\Gamma \vdash \neg A, \Delta}$

1) $\dfrac{\Gamma \vdash \Delta}{\Gamma, 1 \vdash \Delta}$   $\Gamma \vdash 1, \Delta$

0)   $\Gamma, 0 \vdash \Delta$

**Distributivity:**

$\otimes$ – & distr)     $(A \otimes C) \& (B \otimes C) \vdash (A \& B) \otimes C$

**Numerical rules:**

S')   $\beta \vdash \alpha$   for any $\beta \leq \alpha$

$\otimes$ def)   $\alpha \otimes \beta \dashv\vdash \gamma$   where $\gamma = (\alpha + \beta - 1) \vee 0$

$\neg$def)   $\neg \alpha \vdash \gamma$   where $\gamma = 1 - \alpha$

Notes:

1. Absorption is a weak form of contraction for $\mathcal{L}_1$-formulae; consider, in fact, the following derivation, where $L \in \mathcal{L}_1$:

$$\dfrac{\dfrac{L \vdash L}{L \vdash L \otimes L} \text{ abs)} \quad \dfrac{L, L \vdash B}{L \otimes L \vdash B} \otimes L)}{L \vdash B} \text{ cut)}$$

By the way, the rule abs) is stronger than contraction on $\mathcal{L}_1$-formulae since, in exactly the same way, we also have the following (which is very much akin to Lehman's cumulative cut):

$$\dfrac{A \vdash L \quad A, L \vdash B}{A \vdash B}$$

2. Rule $\otimes$-& distr) does not hold in linear logic, since it is specific to the possibility function semantics.

3. If we omit numerical rules, the calculus deals with possibilistic logic with an arbitrary product among possibility functions whose unit is **1**, with the only restriction (due to weakening) that $\pi_1 \times \pi_2 \leq \pi_1 \wedge \pi_2$. Any involution $\sim$ such that $\pi \times \sim \pi = \mathbf{0}$ works as negation. Numerical rules force the times operator to represent the Łukasiewicz product, and the negation the corresponding fuzzy complementation.

4. It should be noted that from absorption) and S') it is possible to derive for any $A$ and $B$ in $\mathcal{L}_1$ and for any $\beta \leq \alpha$ the following sequents that strictly correspond to the rules GMP) and S) explicitly stated in [Dubois, Lang and Prade 94]:

   MP)  $\quad A \,\&\, (A \to B) \vdash B$
   GMP) $\quad (\alpha \to A) \,\&\, (\beta \to (A \to B)) \vdash (\alpha \& \beta) \to B$
   S)   $\quad \alpha \to A \vdash \beta \to A$

5. Also the following two sequents, which will turn out useful, can be derived as proved in lemma 3 ($L$ and $M$ are in $\mathcal{L}_1$)):

   $\otimes$red) $\quad (\alpha \to L) \otimes (\beta \to M) \dashv\vdash \phi$
   $\neg$red) $\quad \neg(\alpha \to L) \dashv\vdash \alpha \,\&\, \neg L$

   where $\phi = (\beta \to (L \to M)) \,\&\, (\alpha \to (M \to L)) \,\&\, ((\alpha \,\mathrlap{\mathscr{D}}{} \, \beta) \to (L \oplus M))$.

6. We remind the reader that distributivity of $\otimes$ with respect to $\oplus$, i.e. the sequent $(A \oplus B) \otimes C \dashv\vdash (A \otimes C) \oplus (B \otimes C)$ holds by the logical rules; we shall refer to it as $\otimes$-$\oplus$ distr). In fact, also the right-to-left direction of $\otimes$-& distr) can be obtained from the logical rules. Similarly, $\neg$ def) rule works also in the right-to-left direction.

7. The rules weakening), $\oplus$), $\mathscr{D}$) and $\to$) can be dropped in a minimal presentation.

Soundness of this calculus is easily proved by induction on the proof length; to prove completeness we use this theorem, which provides a normal form to the formulae of the language DPL:

**Theorem 4** *Any formula $A$ is provably equivalent in the calculus DPL to an &-formula, i.e. a formula $A' = \&_{i \in I}(\alpha_i \to L_i)$ where $L_i$ are $\mathcal{L}_1$-formulae.*

It should be noted that this theorem (whose proof is constructive) guarantees that there is a translation of our language into standard possibilistic logic (and vice-versa), since the formula $\&_{i \in I}(\alpha_i \to L_i)$ can be thought of as the equivalent of $\{(L_i, \alpha_i) : i \in I\}$ in the language of Dubois and colleagues.

Moreover, the presence of $\otimes$ endows DPL with a dynamic dimension: assume you are given information tokens from distinct sources; to merge them, you simply connect the tokens by $\otimes$. Reduction of a formula to the normal form (the &-formula) can be seen as the effective process of merging information. The fact that the reduction process is not so trivial (see the proof of theorem 4) makes it clear that there is some work to do for the combination of information, and this work is automatically performed by the proof system of DPL. It may be worth observing that reducibility of DPL formulae seems to be a very fortunate circumstance due to the simplicity of possibility theory; there is not guarantee, in general, that this process can be performed on logics based on more complex measures, like belief functions. We state then the main theorem:

**Theorem 5** *The DPL calculus is sound and complete with respect to the given semantics, i.e., for any closed multiset $\Gamma$ and $\Delta$, the sequent $\Gamma \vdash \Delta$ is proved iff $\Gamma \models \Delta$.*

## 3  THE PREDICATIVE SYSTEM

### 3.1  THE SEMANTICS

We enrich the propositional language with $\forall$-formulae. $\mathcal{C}$ is the set of individual constants and $\mathcal{R}$ that of predicate symbols; an atomic formula has the form $R(t_1, ..., t_n)$, where $t_i$ are either individual constants or variables. $\mathcal{L}_0$ is the set of atomic formulae.

formula ::= atomic_formula $|$ $\alpha$ $|$ $\neg$ formula $|$ formula & formula $|$ formula $\otimes$ formula $|$ $\forall x$ formula

As before, we take $\mathcal{L}$ to be the set of formulae and $\mathcal{L}_1$ the set of formulae with no occurrences of $\alpha$ constants. Symbols $\to$ and $\oplus$ are defined as in the previous section; we introduce: $\exists x A(x) =_{\text{def}} \neg \forall x \neg A(x)$

In the predicative system we took a general modal semantics and then showed that it was possible to consider just a canonical model (theorem 3) without loss of generality. This time we will not introduce a general modal predicative semantics, since it would be very complex. We consider from the beginning just the canonical models, and define validity with respect to them. Let $\mathcal{M}(D)$ denote the set of classical first order models for the language $\mathcal{L}_1$ on the domain $D$; each element $w \in \mathcal{M}(D)$ has the form $\langle D, F_C, F_R \rangle$ where $F_C$ and $F_R$ are the interpretations of individual constants and relation symbols. $\mathcal{P}(D)$ denote the set of possibility distributions from $\mathcal{M}(\mathcal{D})$ to the real interval $[0,1]$; $\mathcal{P}(D)$ is a particular choice for the set $W$ in section 2, so we make use of definition 1 and theorem 1. It maintains the structure of phase space as in the propositional case, and we can build on it the quantal $\mathcal{Q}(D) = \{G \subseteq \mathcal{P}(D) : G = G^{\perp\perp}\}$.

Let $\sigma$ be an arbitrary assignment for the variables on $D$; by $\sigma[x/u]$ we mean the function which differs from $\sigma$ only on $x$, which is mapped to $u$. We define, for any $\mathcal{L}_1$-formula $A$:

$$Mod_{D,\sigma}(A) =_{\text{def}} \{w \in \mathcal{M}(D) : w, \sigma \models A\}$$

Again, for a fixed domain $D$ and an assignment $\sigma$, we define a function $\|\cdot\|_{D,\sigma} : \mathcal{L} \to 2^{\mathcal{P}(D)}$ as follows:

$$\|R(\bar{t})\|_{D,\sigma} = \{\pi : Nec_\pi(Mod_{D,\sigma}(R(\bar{t}))) = 1\}$$

$$\begin{aligned}
\|\alpha\|_{D,\sigma} &= \downarrow \alpha \\
\|\neg A\|_{D,\sigma} &= \|A\|_{D,\sigma}^{\perp} \\
\|A \& B\|_{D,\sigma} &= \|A\|_{D,\sigma} \cap \|B\|_{D,\sigma} \\
\|A \otimes B\|_{D,\sigma} &= \|A\|_{D,\sigma} \times \|B\|_{D,\sigma} \\
\|\forall x A(x)\|_{D,\sigma} &= \bigcap_{u \in D} \|A(x)\|_{D,\sigma[x/u]}
\end{aligned}$$

It is still true that:

$$\begin{aligned}
\|A \to B\|_{D,\sigma} &= \|A\|_{D,\sigma} \Rightarrow \|B\|_{D,\sigma} \\
\|A \oplus B\|_{D,\sigma} &= \|A\|_{D,\sigma} \vee \|B\|_{D,\sigma} \\
\|A \invamp B\|_{D,\sigma} &= (\|A\|_{D,\sigma} + \|B\|_{D,\sigma}) \wedge \mathbf{1} \\
\|\exists x A(x)\|_{D,\sigma} &= \bigcup_{u \in D} \|A(x)\|_{D,\sigma[x/u]} \\
\|L\|_{D,\sigma} &= \{\pi : Nec_\pi(Mod_{D,\sigma}(L)) = 1\} \\
\|\alpha \to L\|_{D,\sigma} &= \{\pi : Nec_\pi(Mod_{D,\sigma}(L)) \geq \alpha\}
\end{aligned}$$

where $\bar{t} = <t_1, ..., t_n>$, $\pi \in \mathcal{P}(D)$ and $L \in \mathcal{L}_1$.

A possibilistic model model is a couple $K = \langle D, \pi \rangle$. We say that $K \models A$ iff $\pi \in \|A\|_{D,\sigma}$ for any assignment $\sigma$. Notice that, if $A$ is a closed formula, then $\|A\|_{D,\sigma}$ does not change for any choice of $\sigma$; so for a closed formula $A$ we let $\|A\|_D = \|A\|_{D,\sigma}$. We define, as in the propositional case:

$$U_{D,\sigma}(A) =_{\text{def}} \bigvee_{\pi \in \|A\|_{D,\sigma}} \pi$$

**Theorem 6** *Given a frame $F$ and an assignment $\sigma$, $\|A\|_{D,\sigma} = \|A\|_{D,\sigma}^{\perp\perp} = \downarrow U_{D,\sigma}(A)$. Moreover, the following statements hold:*

1. $U_{D,\sigma}(\alpha) = \alpha$

2. $U_{D,\sigma}(L) = \lambda w . \begin{cases} 1 & \text{if } w \in Mod_{D,\sigma}(L) \\ 0 & \text{otherwise} \end{cases}$ for any $\mathcal{L}_1$-formula $L$

3. $U_{D,\sigma}(\neg A) =\sim U_{D,\sigma}(A) =_{\text{def}} 1 - U_{D,\sigma}(A)$

4. $U_{D,\sigma}(A \& B) = U_{D,\sigma}(A) \wedge U_{D,\sigma}(B)$

5. $U_{D,\sigma}(A \otimes B) = U_{D,\sigma}(A) \times U_{D,\sigma}(B) = (U_{D,\sigma}(A) + U_{D,\sigma}(B) - \mathbf{1}) \vee \mathbf{0}$

6. $U_{D,\sigma}(\forall x A(x)) = \bigwedge_{u \in D} U_{D,\sigma[x/u]}(A(x))$

7. $U_{D,\sigma}(A \oplus B) = U_{D,\sigma}(A) \vee U_{D,\sigma}(B)$

8. $U_{D,\sigma}(A \invamp B) = (U_{D,\sigma}(A) + U_{D,\sigma}(B)) \wedge \mathbf{1}$

9. $U_{D,\sigma}(A \to B) = (1 - U_{D,\sigma}(A) + U_{D,\sigma}(B)) \wedge \mathbf{1}$

10. $U_{D,\sigma}(\exists x A(x)) = \bigvee_{u \in D} U_{D,\sigma[x/u]}(A(x))$

For any closed formula $A$ and $B$ we have:

$A \models B$ iff $U_D(A) \leq U_D(B)$ for any domain $D$.

Notice that semantical entailment is defined with respect to all the domains $D$; this is a difference from the propositional case, in which we only had to refer to the canonical frame $C$. In algebraic terms this means that we have to check validity with respect to a class of quantales, and not only to a specific one.

## 3.2 THE PREDICATIVE PROOF SYSTEM DPL*

We only have to add some rules to the propositional calculus:

**Structural rules**: Unchanged

**Logical rules**: Add the following rules:

$$\forall) \quad \frac{\Gamma, A(t) \vdash \Delta}{\Gamma, \forall x A(x) \vdash \Delta} \qquad \frac{\Gamma \vdash A(x), \Delta}{\Gamma \vdash \forall x A(x), \Delta} \star$$

$$\exists) \quad \frac{\Gamma, A(x) \vdash \Delta}{\Gamma, \exists x A(x) \vdash \Delta} \star \qquad \frac{\Gamma \vdash A(t), \Delta}{\Gamma \vdash \exists x A(x), \Delta}$$

$\star$ if $x$ is not free in $\Gamma$ and $\Delta$

**Distributivity**: Add the following rule:

$\otimes - \forall$ distr)  $\forall x A(x) \otimes C \dashv\vdash \forall x (A(x) \otimes C)$   if $x$ is not free in $C$.

**Numerical rules**: Unchanged

We extend theorem 4 to the predicative case:

**Theorem 7** *Any closed formula $A$ is provably equivalent in the calculus DPL\* to an &-formula, i.e. a formula $A' = \&_{i \in I}(\alpha_i \to L_i)$ where $L_i$ are $\mathcal{L}_1$-formulae.*

To prove validity we only have to check the new rules; the proof of completeness does not change w.r.t. that of the propositional system. Eventually, we have:

**Theorem 8** *The DPL\* calculus is sound and complete with respect to the given semantics, i.e., for any closed multiset of formulae $\Gamma$ and $\Delta$, the sequent $\Gamma \vdash \Delta$ is proved iff $\Gamma \models \Delta$.*

## 4 PROOFS OF THEOREMS

All the omitted proofs can be found in [Boldrin and Sossai 95].

**Lemma 1** *If $\pi_1 \in \|A\|_F$ and $\pi_2 \leq \pi_1$, then also $\pi_2 \in \|A\|_F$.*

**Proof** By induction on the complexity of $A$:

$A = \alpha$ :  $\pi_1 \in \|\alpha\|_F = \downarrow \alpha$ implies $\pi_1 \leq \alpha$. Then $\pi_2 \leq \pi_1 \leq \alpha$, and so $\pi_2 \in \|\alpha\|_F$.

$A = p$ :  Since $\pi_1 \in \|p\|_F$, we have $\bigvee_{w \notin V(p)} \pi_1(w) = 0$. Since $\pi_2 \leq \pi_1$, $\bigvee_{w \notin V(p)} \pi_2(w) \leq \bigvee_{w \notin V(p)} \pi_1(w) = 0$, hence $\pi_2 \in \|p\|_F$.

$A = \neg B$ :  Since $\pi_1 \in \|A\|_F = \|B\|_F^{\perp}$, we have that for any $\sigma \in \|B\|_F$, $\pi_1 \times \sigma = \mathbf{0}$. Then for any $\sigma \in \|B\|_F$, $\pi_2 \times \sigma \leq \pi_1 \times \sigma = \mathbf{0}$, and so $\pi_2 \in \|B\|_F^{\perp}$.

$A = C \,\&\, B$: Since $\pi_1 \in \|C \wedge B\|_F$, by def. $\pi_1 \in \|C\|_F$ and $\pi_1 \in \|B\|_F$. By inductive hypothesis $\pi_2 \in \|C\|_F$ and $\pi_2 \in \|B\|_F$, and then $\pi_2 \in \|C \wedge B\|_F$.

$A = C \otimes B$: Let $\pi_1^1 \in \|C\|_F$, $\pi_1^2 \in \|B\|_F$ and $\pi_1 = \pi_1^1 \times \pi_1^2$. Take $\pi_2^2 = \pi_1^2$, and $\pi_2^1$ defined as:

$$\pi_2^1(w) = \begin{cases} 1 - \pi_1^2(w) + \pi_2(w) & \text{if } \pi_2(w) \neq 0 \\ 0 & \text{otherwise} \end{cases}$$

We have then $\pi_2^1 \in \|C\|_F$ by inductive hyp., since $\pi_2^1 \leq \pi_1^1$ (easy to see) and $\pi_2^2 \in \|B\|_F$. Moreover, $\pi_2^1 \times \pi_2^2 = \pi_2$. Hence we have proved that $\pi_2 \in \|C \otimes B\|_F$.

so the proof is over. □

**Proof of Theorem 1**

1. ⊇) Take $\pi$ such that $\pi \times \pi_G = \mathbf{0}$; then, for any $\sigma \in G$, $\pi \times \sigma \leq \pi \times \pi_G = \mathbf{0}$, so $\pi \in G^\perp$.

   ⊆) Take $\pi$ such that $\pi \times \pi_G \neq \mathbf{0}$; then there is a $w_0$ such that $\pi(w_0)\pi_G(w_0) > 0$. Since $\pi_G(w_0) = \bigvee_{\sigma \in G} \sigma(w_0)$, there exists a $\sigma \in G$ with $\sigma(w_0) > 0$; for this $\sigma$, $\pi \times \sigma > \mathbf{0}$, so $\pi \notin G^\perp$.

2. ⊇) Take $\pi \in \downarrow \pi_G$; then $\pi \leq \pi_G$, hence for any $\sigma$ such that $\sigma \times \pi_G = \mathbf{0}$, we have: $\sigma \times \pi \leq \sigma \times \pi_G = \mathbf{0}$

   ⊆) Take $\pi \notin \downarrow \pi_G$; then $\pi \not\leq \pi_G$, so there exists a $w_0$ such that $\pi(w_0) > \pi_G(w_0)$. Take now $\sigma$ defined as follows: $\sigma(w) = 1 - \pi_G(w)$; clearly, it is $\sigma \times \pi_G = \mathbf{0}$, which implies $\sigma \in G^\perp$ by the first point; it is also $\sigma(w_0) = 1 - \pi_G(w_0) > 0$ since $\pi_G(w_0) < 1$, but $\sigma(w_0)\pi(w_0) > \sigma(w_0)\pi_G(w_0)\pi_G = \mathbf{0}$, so $\sigma \times \pi > \mathbf{0}$, and $\pi \notin G^{\perp\perp}$

so the proof is over. □

**Proof of theorem 2** Let $U_F(A) = \bigvee \|A\|_F$; we prove that $\|A\|_F = \downarrow U_F(A)$, since we know from theorem 1 that $\|A\|_F^{\perp\perp} = \downarrow \bigvee \|A\|_F$.

⊆ ) is obvious.
⊇ ) We prove by induction that $U_F(A) \in \|A\|_F$; then, by lemma 1, we know that any $\pi$ so that $\pi \leq U_F(A)$ is in $\|A\|_F$. (We omit the subscripts $F$):

$A = \alpha$: Clearly, $U_F(\alpha) = \alpha$, in fact: for any $\pi \in \|\alpha\|$, $\pi \leq \alpha$; and $\alpha \in \|\alpha\|$.

$A = L$ $\mathcal{L}_1$-formula: We show that $U_F(L)(w) = \begin{cases} 1 & \text{if } w \in V(L) \\ 0 & \text{if } w \notin V(L) \end{cases}$

- For any $\pi \in \|L\| = \{\pi : Nec_\pi(V(L)) = 1\}$ it must be the case that $\bigvee_{w \notin V(L)} \pi(w) = 0$. This means that for any $w \notin V(L)$, $\pi(w) = 0$; for $w \in V(L)$, it is certainly $\pi(w) \leq 1$.

- The characteristic function of $V(L)$ stands in $\|L\|$ (easy to verify).

$A = \neg B$: We show that $U(\neg B) = \mathbf{1} - U(B)$:

- For any $\pi \in \|\neg B\|$, it is (theorem 1) $\pi \times U(B) = \mathbf{0} \vee (\pi + U(B) - \mathbf{1}) = \mathbf{0}$, and so $\pi + U(B) - \mathbf{1} \leq \mathbf{0}$. which implies $\pi \leq \mathbf{1} - U(B)$.
- $\mathbf{1} - U(B) \in \|\neg B\|$, since $U(B) \times (\mathbf{1} - U(B)) = \mathbf{0}$.

$A = B \,\&\, C$: We show that $U(B \,\&\, C) = U(B) \wedge U(C)$:

- Take $\pi \in \|B\&C\|$; by definition, $\pi \in \|B\|$ and $\pi \in \|C\|$. So it is $\pi \leq U(B)$ and $\pi \leq U(C)$ and, eventually, $\pi \leq U(B) \wedge U(C)$.
- $U(B) \wedge U(C) \leq U(B)$ and $U(B) \wedge U(C) \leq U(C)$; by inductive hyp. $U(B) \in \|B\|$ and $U(C) \in \|C\|$ so, by lemma 1, $U(B) \wedge U(C) \in \|B\|$ and $U(B) \wedge U(C) \in \|C\|$. So, by definition, $U(B) \wedge U(C) \in \|B \,\&\, C\|$.

$A = B \otimes C$: We show that $U(B \otimes C) = U(B) \times U(C)$:

- Take $\pi \in \|B \otimes C\|$; then there exist $\pi_1 \in \|B\|$ and $\pi_2 \in \|C\|$ so that $\pi = \pi_1 \times \pi_2$. Then $\pi_1 \leq U(B)$ and $\pi_2 \leq U(C)$. By monotonicity of $\times$, $\pi \leq U(B) \times U(C)$.
- $U(B) \times U(C) \in \|B \otimes C\|$, since it is the product of two functions which (by induction hyp.) stand respectively in $\|B\|$ and $\|C\|$.

The proof by induction is over; using definitions we can also calculate:

$$U(B \oplus C) = U(\neg(\neg B \,\&\, \neg C)) = U(B) \vee U(C).$$

$$U(B \,⅋\, C) = U(\neg(\neg B \otimes \neg C)) = (U(B) + U(C)) \wedge \mathbf{1}.$$

$$U(B \to C) = U(\neg(B \otimes \neg C)) = (\mathbf{1} - U(B) + U(C)) \wedge \mathbf{1}.$$

so the proof is over. □

**Lemma 2** *Let $B \dashv\vdash \&_{i \in I}(\beta_i \to L_i)$ where $L_i$ are $\mathcal{L}_1$-formulae, and, for any $J \subseteq I$, $\alpha_J = \neg \bigoplus_{j \notin J} \beta_j$ and $M_J = \neg \left( \&_{j \in J} L_j \,\&\, \&_{j \notin J} \neg L_j \right)$ ($M_J \in \mathcal{L}_1$). Then the following derivation holds:*

$$\neg B \dashv\vdash \&_{J \subseteq I}(\alpha_J \to M_J)$$

**Lemma 3** *The following sequent can be obtained in DPL for $L$ and $M$ in $\mathcal{L}_1$:*

$$\begin{array}{l} (\beta \to (L \to M)) \,\& \\ (\alpha \to (M \to L)) \,\& \\ ((\alpha \,⅋\, \beta) \to (L \oplus M)) \end{array} \vdash \begin{array}{l} ((\alpha \to (M \to L)) \otimes (\beta \to (L \to M))) \,\& \\ ((\alpha \to (M \to L)) \otimes (\beta \to (L \oplus M))) \,\& \\ ((\alpha \to (M \oplus L)) \otimes (\beta \to (L \to M))) \,\& \\ ((\alpha \to (M \oplus L)) \otimes (\beta \to (L \oplus M))) \end{array}$$

**Lemma 4** *The following equivalence can be proved in DPL for any $\mathcal{L}_1$-formulae $L$ and $M$:*

$\otimes$ red) $(\alpha \to L) \otimes (\beta \to M) \dashv\vdash (\beta \to (L \to M)) \,\&\, (\alpha \to (M \to L)) \,\&\, ((\alpha \,⅋\, \beta) \to (L \oplus M))$

**Lemma 5** *The following equivalence can be proved in DPL for any $\mathcal{L}_1$-formula L:*

$$\neg\ red)\quad \neg(\alpha \to L) \dashv\vdash \alpha\ \&\ \neg L$$

**Proof of theorem 5 (soundness and completeness)** Soundness is easy to prove, by induction on the derivation; as a matter of fact, logical and structural rules (apart of weakening and absorption) are sound because the interpretation structure is a Girard quantal (see [Girard 87]), with the proviso that **1** coincides with $\top$. For the others the proof is straightforward:

**weakening)**: Assume $U(\bigotimes \Delta) \leq U(B)$; then $U(\bigotimes \Delta \otimes A) = U(\bigotimes \Delta) \times U(A) \leq U(\bigotimes \Delta) \leq U(B)$ because of monotonicity of $\times$.

**absorption)**: We prove that if $U(A) \leq U(L)$ then $U(A) \times U(L) = U(A)$. By theorem 2 $U(L)(w) \in \{0,1\}$, hence we have $(U(A) \times U(L))(w) = (U(A)(w) + U(L)(w) - 1) \vee 0 =$
$\begin{cases} U(A)(w) & \text{if } U(L)(w) = 1 \\ 0 & \text{if } U(L)(w) = 0 \end{cases}$ hence $U(A)(w) \wedge U(L)(w) = U(A)(w)$

**⊗-& distr)**: $U((A\ \&\ B) \otimes C) = U(A\ \&\ B) \times U(C) = (U(A\ \&\ B) + U(C) - 1) \vee \mathbf{0} = ((U(A) \wedge U(B)) + U(C) - 1) \vee \mathbf{0} = ((U(A) + U(C) - 1) \wedge (U(B) + U(C) - 1)) \vee \mathbf{0} = ((U(A) + U(C) - 1) \vee \mathbf{0}) \wedge ((U(B) + U(C) - 1) \vee \mathbf{0}) = (U(A) \times U(C)) \wedge (U(B) \times U(C)) = U(A \otimes C) \wedge U(B \otimes C) = U((A \otimes C)\ \&\ (B \otimes C))$

**S')**: $U(\beta) = \beta \leq \alpha = U(\alpha)$

**⊗ def)**: $U(\alpha \otimes \beta) = U(\alpha) \times U(\beta) = \alpha \times \beta = \gamma$ where $\gamma = (\alpha + \beta - 1) \vee 0$

**¬ def)**: $U(\neg\alpha) = 1 - U(\alpha) = 1 - \alpha = \gamma$ where $\gamma = 1 - \alpha$.

Let us now come to completeness; assume $U_F(A) \leq U_F(B)$; using theorem 4 we have $A \dashv\vdash \&_{i \in I}(\alpha_i \to L_i)$ and $B \dashv\vdash \&_{j \in J}(\beta_j \to M_j)$. Our hypothesis can be equivalently formulated as $U(\&_{i \in I}(\alpha_i \to L_i)) \leq U((\beta_j \to M_j))$ for any $j \in J$. We prove by induction on the number of elements in $I$ that, under this hypothesis, $\&_{i \in I}(\alpha_i \to L_i) \vdash (\beta_j \to M_j)$.

$|I| = 1$ : We have then $U(\alpha \to L) \leq U(\beta \to M)$. There are just two cases in which this can happen:

- $V(L) \subseteq V(M)$ and $\beta \leq \alpha$. Then by classical completeness we have $L \vdash M$, and by rule S') we have $\beta \vdash \alpha$; hence:

$$\frac{\beta \vdash \alpha \quad L \vdash M}{\dfrac{\alpha \to L, \beta \vdash M}{\alpha \to L \vdash \beta \to M}}$$

- $\beta = 0$. In this case $\dfrac{\dfrac{\mathbf{0} \vdash M}{\vdash \mathbf{0} \to M}}{\alpha \to L \vdash \mathbf{0} \to M}$

$|I| = n$ : We have $U(\&_{i \leq n-1}(\alpha_i \to L_i)\ \&\ (\alpha_n \to L_n)) \leq U(\beta \to M)$. Using theorem 2 we have:

$$U(\&_{i \leq n-1}(\alpha_i \to L_i)\ \&\ (\alpha_n \to L_n))(w) =$$
$$\begin{cases} U(\&_{i \leq n-1}(\alpha_i \to L_i))(w) & \text{if } w \in V(L_n) \\ (1-\alpha_n) \wedge U(\&_{i \leq n-1}(\alpha_i \to L_i))(w) & \text{otherwise} \end{cases}$$
$$U(\beta \to M) = \begin{cases} 1 & \text{if } w \in V(M) \\ 1 - \beta & \text{otherwise} \end{cases}$$

It follows that there are two possibilities:

- $U(\&_{i \leq n-1}(\alpha_i \to L_i))(w) \leq 1 - \beta$ for any $w$ such that $w \notin V(L_n)$ and $w \notin V(M)$. In this case it happens that $U(\&_{i \leq n-1}(\alpha_i \to L_i)) \leq U(\beta \to M)$. Then we can make the simple derivation

$$\frac{\&_{i \leq n-1}(\alpha_i \to L_i) \vdash \beta \to M}{\&_{i \leq n-1}(\alpha_i \to L_i)\ \&\ (\alpha_n \to L_n) \vdash \beta \to M}$$

- If the first possibility does not hold, then it must be the case that $1 - \alpha_n \leq 1 - \beta$, i.e. $\beta \leq \alpha_n$; we also have (this is true in any case): $U(\&_{i \leq n-1}(\alpha_i \to L_i)) \leq U(\beta \to (M \oplus \neg L_n))$. By inductive hypothesis we have $\&_{i \leq n-1}(\alpha_i \to L_i) \vdash \beta \to (M \oplus \neg L_n)$
Then we have the following (notice that from $\beta \leq \alpha$ we have $\beta \vdash \alpha_n$, hence $\alpha_n \to L_n \vdash \beta \to L_n$):

$$\frac{\dfrac{\&_{i \leq n-1}(\alpha_i \to L_i) \vdash \beta \to (M \oplus \neg L_n) \quad \alpha_n \to L_n \vdash \beta \to L_n}{\dfrac{\&_{i \leq n-1}(\alpha_i \to L_i)\ \&\ (\alpha_n \to L_n) \vdash (\beta \to (M \oplus \neg L_n))\ \&\ (\beta \to L_n)}{\dfrac{\&_{i \leq n}(\alpha_i \to L_i) \vdash \beta \to ((M \oplus \neg L_n)\ \&\ L_n)}{\&_{i \leq n}(\alpha_i \to L_i) \vdash \beta \to M}}}$$

From $\&_{i \in I}(\alpha_i \to L_i) \vdash (\beta_j \to M_j)$ for any $j \in J$ we get $\&_{i \in I}(\alpha_i \to L_i) \vdash \&_{j \in J}(\beta_j \to M_j)$, hence $L \vdash M$. □

**Lemma 6** *(predicative homologue of lemma 1). If $\pi_1 \in \|A\|_{D,\sigma}$ and $\pi_2 \leq \pi_1$, then also $\pi_2 \in \|A\|_{D,\sigma}$.*

**Proof** By induction on the complexity of $A$. The only case which must be considered is $A = \forall x B(x)$, since all the other cases do not change with respect to the propositional proof (lemma 1).

$A = \forall x B(x)$: Since $\pi_1 \in \|\forall x A(x)\|_{D,\sigma} = \bigwedge_{u \in D} \|A(x)\|_{D,\sigma[x/u]}$, it is $\pi_1 \in \|A(x)\|_{D,\sigma[x/u]}$ for all $u \in D$, and so (ind. hyp.) $\pi_2 \in \|A(x)\|_{D,\sigma[x/u]}$ for all $u \in D$, hence $\pi_2 \in \bigwedge_{u \in D} \|A(x)\|_{D,\sigma[x/u]} = \|\forall x A(x)\|_{D,\sigma}$. □

**Proof of theorem 6** (predicative homologue of theorem 2). Let $U_{D,\sigma}(A) = \bigvee \|A\|_{D,\sigma}$; we prove that $\|A\|_{D,\sigma} = \downarrow U_{D,\sigma}(A)$, since we know from theorem 1 that $\|A\|_{D,\sigma}^{\perp\perp} = \downarrow \bigvee \|A\|_{D,\sigma}$.

$\subseteq$ ) is obvious.
$\supseteq$ ) We prove by induction that $U_{D,\sigma}(A) \in \|A\|_{D,\sigma}$; then, by lemma 4, we know that any $\pi$ so that $\pi \leq U_{D,\sigma}(A)$ is in $\|A\|_{D,\sigma}$. The only case which must be considered is $A = \forall x B(x)$, since all the others do not change w.r.t. the propositional case.

We show that $U(\forall x B(x) = \bigwedge_{u \in D} U_{D,\sigma[x/u]}(B(x))$:

- Take $\pi \in \|\forall x B(x)\| = \bigwedge_{u \in D} \|B(x)\|_{D,\sigma[x/u]}$; then $\pi \leq U_{D,\sigma[x/u]}(B(x))$ for any $u \in D$, hence $\pi \leq \bigwedge_{u \in D} U_{D,\sigma[x/u]}(B(x))$.

- $\bigwedge_{u \in D} U_{D,\sigma[x/u]}(B(x)) \leq U_{D,\sigma[x/u]}(B(x))$ for any $u \in D$. By ind. hyp, $U_{D,\sigma[x/u]}(B(x)) \in \|(B(x))\|_{D,\sigma[x/u]}$ for any $u \in D$; by lemma 6, $\bigwedge_{u \in D} U_{D,\sigma[x/u]}(B(x)) \in \|(B(x))\|_{D,\sigma[x/u]}$ for any $u \in D$, and so $\bigwedge_{u \in D} U_{D,\sigma[x/u]}(B(x)) \in \bigcap_{u \in D} \|(B(x))\|_{D,\sigma[x/u]} = \|\forall x B(x)\|$.

so the proof is over.   □

**Proof of theorem 8** (predicative homologue of theorem 5). To prove soundness we only have to check the new rules for validity; the $\forall$-rules are in fact valid because of validity of linear logic. The only rule which has to be verified is:

$\otimes - \forall$ distr)    $\forall x A(x) \otimes C \dashv\vdash \forall x(A(x) \otimes C)$   if $x$ is not free in $C$.

For any $D$: $U_D(\forall x A(x) \otimes C) = U_D(\forall x A(x)) \times U_D(C) = \bigwedge_{u \in D} U_{D,[x/u]}(A(x)) \times U_D(C) = \bigwedge_{u \in D}(U_{D,[x/u]}(A(x)) \times U_{D,[x/u]}(C)) = \bigwedge_{u \in D}(U_{D,[x/u]}(A(x) \otimes C)) = U_D(\forall x(A(x) \otimes C))$

Completeness is proved exactly as completeness of the propositional system, where references to theorem 4 are substituted by references to theorem 7.   □

## Acknowledgments

We are grateful to Alessandro Saffiotti for several precious advices, and to Nino Trainito for a careful reading of the paper.